\pdfoutput=1

\documentclass[11pt]{article}

\usepackage[]{ACL2023}

\usepackage{times}
\usepackage{latexsym}
\usepackage{booktabs}
\usepackage{graphicx}
\usepackage{multirow}
\usepackage{times}
\usepackage{amsfonts,amsmath,amssymb,bm}
\usepackage{graphicx}
\usepackage{xcolor}
\usepackage{setspace}
\usepackage{pdflscape}
\usepackage{pdfpages}
\usepackage{enumitem}
\usepackage{rotating}
\usepackage{caption}
\usepackage{subcaption}
\usepackage{pythonhighlight}

\lstnewenvironment{mypython}[1][]{\lstset{style=mypython,#1}}{}

\usepackage[english]{babel}
\usepackage{hyperref}
\addto\captionsenglish{%
}
\addto\extrasenglish{%
}

\usepackage[T1]{fontenc}

\usepackage[utf8]{inputenc}

\usepackage{microtype}

\usepackage{inconsolata}

\title{A Practical Toolkit for Multilingual Question and Answer Generation}

\author{Asahi Ushio \and Fernando Alva-Manchego \and Jose Camacho-Collados\\
  Cardiff NLP, School of Computer Science and Informatics, Cardiff University, UK\\
  \texttt{\{UshioA,AlvaManchegoF,CamachoColladosJ\}@cardiff.ac.uk}
  \\}

\begin{document}
\maketitle
\begin{abstract}
Generating questions along with associated answers from a text has applications in several domains, such as creating reading comprehension tests for students, or improving document search by providing auxiliary questions and answers based on the query. Training models for question and answer generation (QAG) is not straightforward due to the expected structured output (i.e.\ a list of question and answer pairs), as it requires more than generating a single sentence. This results in a small number of publicly accessible QAG models. In this paper, we introduce AutoQG, an online service for multilingual QAG, along with \texttt{lmqg}, an all-in-one Python package for model fine-tuning, generation, and evaluation. We also release QAG models in eight languages fine-tuned on a few variants of pre-trained encoder-decoder language models, which can be used online via AutoQG or locally via \texttt{lmqg}. With these resources, practitioners of any level can benefit from a toolkit that includes a web interface for end users, and easy-to-use code for developers who require custom models or fine-grained controls for generation.

\end{abstract}

\section{Introduction}

Question and answer generation (QAG) is a text generation task seeking to output a list of question-answer pairs based on a given paragraph or sentence (i.e. the context). 
It has been used in many NLP applications, including unsupervised question answering modeling \cite{lewis-etal-2019-unsupervised,zhang-bansal-2019-addressing,puri-etal-2020-training}, fact-checking \cite{ousidhoum-etal-2022-varifocal}, semantic role labeling \cite{pyatkin-etal-2021-asking}, and as an educational tool
\cite{heilman-smith-2010-good,lindberg-etal-2013-generating}. The most analysed setting in the literature, however, has been question generation (QG) with pre-defined answers, as this simplifies the task and makes the evaluation more straightforward.

\begin{figure}[t!]
 \centering
 \includegraphics[width=1.0\columnwidth]{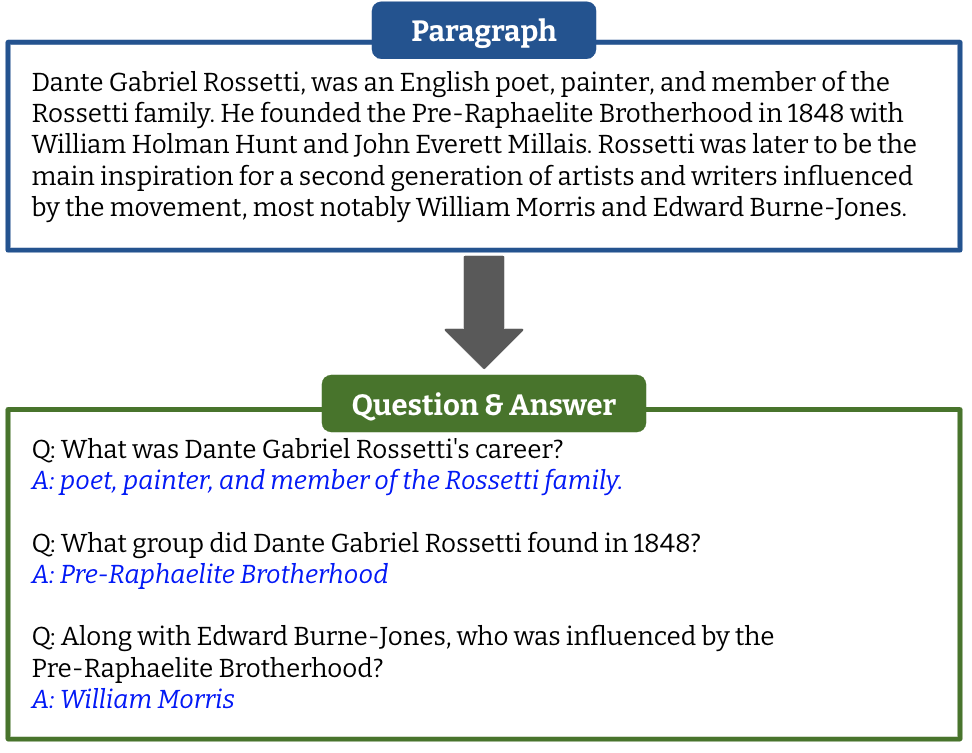}
\caption{An example of question and answer generation given a paragraph as context.}
 \label{fig:overview}
\end{figure}

Despite its versatility, QAG remains a challenging task due to the difficulty of generating compositional outputs containing a list of question and answer pairs as shown in \autoref{fig:overview}, with recent works mainly relying on extended pipelines that include several ad-hoc models \cite{lewis-etal-2021-paq,bartolo-etal-2021-improving}. These works integrate QAG into their in-house software, preventing models to be publicly released, and their complex pipelines make them hard to reproduce and use by practitioners.

In this paper, we introduce an open set of software tools and resources to assist on the development and employment of QAG models for different types of users. We publicly release the following resources:\footnote{All the resources except for the datasets are released under an open MIT license, while the datasets follow the license of their original release.}
\begin{itemize}
    \item \texttt{lmqg},\footnote{\url{https://github.com/asahi417/lm-question-generation}} a Python package for QAG model fine-tuning and inference on encoder-decoder language models (LMs), as well as evaluation scripts, and a deployment API hosting QAG models for developers;
    \item 16 models for English, and three diverse models for each of the seven languages integrated into our library, all fine-tuned on QG-Bench \cite{ushio-etal-2022-generative} and available on the HuggingFace hub~\cite{wolf-etal-2020-transformers};\footnote{\url{https://huggingface.co/lmqg}} 
    \item \textit{AutoQG} (\url{https://autoqg.net}), a website where developers and end users can interact with our multilingual QAG models.
\end{itemize}




\section{Resources: Models and Datasets}\label{sec:released-models}
Our QAG toolkit makes use of pre-existing models and datasets, fully compatible with the HuggingFace hub. This makes our library easily extendable in the future as newer datasets and better models emerge.
In this section, we describe the datasets (\autoref{sec:data}) and models (\autoref{sec:models}) currently available through \texttt{lmqg} and AutoQG. 

\subsection{Multilingual Datasets}\label{sec:data}
Our toolkit integrates all QG datasets available in QG-Bench \cite{ushio-etal-2022-generative}. QG-Bench is a multilingual QG benchmark consisting of a suite of unified QG datasets in different languages. In particular, we integrate the following datasets: SQuAD (English), 
SQuADShifts \cite{miller2020effect} (English), SubjQA \cite{bjerva-etal-2020-subjqa} (English), JAQuAD \cite{so2022jaquad} (Japanese), GerQuAD \cite{GermanQuAD} (German), SberQuAd \cite{efimov2020sberquad} (Russian), KorQuAD \cite{lim2019korquad1} (Korean), FQuAD \cite{dhoffschmidt-etal-2020-fquad} (French), Spanish SQuAD \cite{2016arXiv160605250R} (Spanish), and Italian SQuAD \cite{squad_it} (Italian).
QG-Bench is available through our official \texttt{lmqg} HuggingFace project page and GitHub\footnote{\url{https://github.com/asahi417/lm-question-generation/blob/master/QG_BENCH.md}}.

\subsection{Models}\label{sec:models}
Aiming to make QAG models publicly accessible in several languages, we used \texttt{lmqg} to fine-tune LMs using QG-Bench (\autoref{sec:data}).
First, we defined a pipeline QAG model architecture consisting of two independent models: one for answer extraction (AE) and one for question generation (QG).
During training, the AE model learns to find an answer in each sentence of a given paragraph, while the QG model learns to generate a question given an answer from a paragraph.
To generate question-answer pairs at generation time, the AE model first extracts answers from all the sentences in a given paragraph, and then these are used by the QG model to generate a question for each answer.
While not directly evaluated in this paper, we also integrated other types of QAG methods such as multitask and end2end QAG \cite{ushio-etal-2023-an-empirical}, all available via the \texttt{lmqg} library (\autoref{sec:lmqg}) as well as AutoQG (\autoref{sec:autoqg}).

As pre-trained LMs, we integrated T5 \cite{T5}, Flan-T5 \cite{chung2022scaling}, and BART \cite{lewis-etal-2020-bart} for English; and
mT5 \cite{xue-etal-2021-mt5} and mBART \cite{liu-etal-2020-multilingual-denoising} for non-English QAG models. 
The pre-trained weights were taken from checkpoints available in the HuggingFace Hub as below:
\begin{itemize}
    \setlength\itemsep{0em}
    \item \texttt{t5-\{small,base,large\}}
    \item \texttt{google/flan-t5-\{small,base,large\}}
    \item \texttt{facebook/bart-\{base,large\}}
    \item \texttt{google/mt5-\{small,base\}}
    \item \texttt{facebook/mbart-large-cc25}
\end{itemize}

All the fine-tuned QAG models are publicly available in our official HuggingFace Hub. While we initially integrated these models, users can easily fine-tune others using \texttt{lmqg}, as we show in \autoref{sec:lmqg}.

\section{\texttt{lmqg}: An All-in-one QAG Toolkit} \label{sec:lmqg}
In this section, we introduce \texttt{lmqg} ({\bf L}anguage {\bf M}odel for {\bf Q}uestion {\bf G}eneration), a Python library for fine-tuning LMs on QAG (\autoref{sec:lmqg-finetuning}), generating question-answer pairs (\autoref{sec:lmqg-generation}), and evaluating QAG models (\autoref{sec:lmqgeval}). Additionally, with \texttt{lmqg}, we build a REST API to host QAG models to generate question and answer interactively (\autoref{sec:autoqg}).
\texttt{lmqg} is inter-operable with the HuggingFace ecosystem, as it can directly make use of the datasets and models already shared on the HuggingFace Hub.

\subsection{QAG Model Fine-tuning 
}\label{sec:lmqg-finetuning}
Fine-tuning is performed via \texttt{GridSearcher}, a class to run encoder-deocoder LM fine-tuning with hyper-parameter optimization (see \autoref{sec:grid-search} for more details).
For example, the following code shows how we can fine-tune T5 \cite{T5} on SQuAD \cite{rajpurkar-etal-2016-squad}, with the QAG model explained in \autoref{sec:models}. Since we decomposed QAG into AE and QG, two models need to be fine-tuned independently.

\begin{mypython}[label=SO-test]
  from lmqg import GridSearcher

  # instantiate AE trainer 
  trainer_ae = GridSearcher(
    dataset_path="lmqg/qg_squad",
    input_types="paragraph_sentence",
    output_types="answer",
    model="t5-large")
  
  # train AE model
  trainer_ae.train()
  
  # instantiate QG trainer 
  trainer_qg = GridSearcher(
    dataset_path="lmqg/qg_squad",
    input_types="paragraph_answer",
    output_types="question",
    model="t5-large")
  
  # train QG model
  trainer_qg.train()
\end{mypython}

The corresponding dataset, \texttt{lmqg/qg\_squad},\footnote{\url{https://huggingface.co/datasets/lmqg/qg_squad}} has as columns: \texttt{paragraph\_answer} (answer-highlighted paragraph), \texttt{paragraph\_sentence} (sentence-highlighted paragraph), \texttt{question} (target question), and \texttt{answer} (target answer).
The input and the output to the QG model are \texttt{paragraph\_answer} and \texttt{question}, while those to the AE model are \texttt{paragraph\_sentence} and \texttt{answer}.
The inputs and the outputs can be specified by passing the name of each column in the dataset to the arguments, \texttt{input\_types} and \texttt{output\_types} when instantiating \texttt{GridSearcher}.

\subsection{QAG Model Generation 
} \label{sec:lmqg-generation}
In order to generate question-answer pairs from a fine-tuned QAG model, \texttt{lmqg} provides the \texttt{TransformersQG} class. It takes as input a path to a local model checkpoint or a model name on the HuggingFace Hub in order to generate predictions in a single line of code.
The following code snippet shows how to generate a list of question and answer pairs with the fine-tuned QAG model presented in \autoref{sec:models}.
\texttt{TransformersQG} decides which model to use for each of AE and QG based on the arguments \texttt{model\_ae} and \texttt{model}.

\begin{mypython}[label=SO-test]
  from lmqg import TransformersQG

  # instantiate model
  model = TransformersQG(
    model="lmqg/t5-base-squad-qg",
    model_ae="lmqg/t5-base-squad-ae"
  )

  # input paragraph
  x = """William Turner was an English
  painter who specialised in watercolour 
  landscapes. One of his best known 
  pictures is a view of the city of 
  Oxford from Hinksey Hill."""

  # generation
  model.generate_qa(x)
  [
   (
    "Who was an English painter 
     specialised in watercolour 
     landscapes?",
    "William Turner"
   ),
   (
    "Where is William Turner's 
     view of Oxford?",
    "Hinksey Hill."
   )
  ]
\end{mypython}

\subsection{QAG Model Evaluation 
} \label{sec:lmqgeval}
Similar to other text-to-text generation tasks, we implement an evaluation mechanism that compares the set of generated question-answer pairs $\mathcal{\tilde{Q}}_p=\{ (\tilde{q}^1, \tilde{a}^1), (\tilde{q}^2, \tilde{a}^2), \dots \}$ to a reference set of gold question-answer pairs $\mathcal{Q}_p=\{ (q^1, a^1), (q^2, a^2), \dots \}$ given an input paragraph $p$.
Let us define a function to evaluate a single question-answer pair to its reference pair as
\begin{align}
    d_{q, a, \tilde{q}, \tilde{a}} &= s\big(t(q,a), t(\tilde{q}, \tilde{a})\big) \label{eq:d}\\
    t(q, a)&=``\texttt{question:}\{q\}, \texttt{answer:}\{a\}\textrm'\textrm' \label{eq:prompt-single-pair}
\end{align}
where $s$ is a reference-based metric,
and we compute the $F_1$ score as the final metric as below:
\begin{align}
    F_1 &= 2 \frac{R\cdot P}{R + P} \label{eq:f1}\\
    R &= {\rm mean}\Big(\Big[
    \max_{(q, a)\in\mathcal{Q}_c}
        \big( d_{q, a, \tilde{q}, \tilde{a}} \big)
    \Big]_{(\tilde{q}, \tilde{a})\in\mathcal{\tilde{Q}}_c}\Big) \label{eq:recall}\\
    P &= {\rm mean}\Big(\Big[
    \max_{(\tilde{q}, \tilde{a})\in\mathcal{\tilde{Q}}_c}
        \big( d_{q, a, \tilde{q}, \tilde{a}} \big)
    \Big]_{(q, a)\in\mathcal{Q}_c}\Big) \label{eq:precision}
\end{align}

Conceptually, the recall \eqref{eq:recall} and precision \eqref{eq:precision} computations attempt to ``align'' each generated question-answer pair to its ``most relevant'' reference pair. As with traditional precision and recall metrics, precision is aimed at evaluating whether the predicted question-answer pairs are \textit{correct} (or in this case, aligned with the reference question-answer pairs), and recall tests whether there are enough high-quality question-answer pairs. Thus, we refer to the score in \eqref{eq:f1} as the \textbf{QAAligned F1 score}.
The quality of the alignment directly depends on the underlying metric $s$.
Furthermore, 
the complexity of QAAligned is no more than the complexity of the underlying metric,
and invariant to the order of generated pairs because of the alignment at computing recall and precision.

Out-of-the-box, \texttt{lmqg} implements two variants based on the choice of \texttt{base\_metric} $s$ (used for evaluation in \autoref{sec:eval}): QAAligned BS using BERTScore \cite{zhang2019bertscore} and QAAligned MS using MoverScore \cite{zhao-etal-2019-moverscore}. We selected these two metrics as they correlate well with human judgements in QG \cite{ushio-etal-2022-generative}.
Nevertheless, the choice of \texttt{base\_metric} is flexible and users can employ other natural language generation (NLG) evaluation metrics such as BLEU4 \cite{papineni-etal-2002-bleu}, METEOR \cite{denkowski-lavie-2014-meteor}, or ROUGE\textsubscript{L} \cite{lin-2004-rouge}.

With \texttt{lmqg}, QAAligned score can be computed with the \texttt{QAAlignedF1Score} class as shown in the code snippet below:

\begin{mypython}[label=SO-test]
  from lmqg import QAAlignedF1Score

  # gold reference and generation
  ref = [
  "question: What makes X?, answer: Y",
  "question: Who made X?, answer: Y"]
  pred = [
  "question: What makes X?, answer: Y",
  "question: Who build X?, answer: Y",
  "question: When X occurs?, answer: Y"]

  # compute QAAligned BS
  scorer = QAAlignedF1Score(
    base_metric="bertscore")
  scorer.get_score(pred, ref)

  # compute QAAligned MS
  scorer = QAAlignedF1Score(
    base_metric="moverscore"
  )
  scorer.get_score(pred, ref)
\end{mypython}

\section{Evaluation}
\label{sec:eval}

We rely on the QAG models and datasets included in the library (see \autoref{sec:released-models}). The individual QG components of each model (i.e.\ the generation of a question given an answer in a paragraph) were extensively evaluated in \citet{ushio-etal-2022-generative}. For this evaluation, therefore, we focus on the quality of the predicted questions and answers given a paragraph (i.e. the specific answer is not pre-defined). For each model, we fine-tune, make predictions and compute their QAAligned scores via \texttt{lmqg}.

\subsection{Results}

\paragraph{Monolingual evaluation (English).} \autoref{tab:qaa-en} presents the test results on SQuAD for seven English models based on BART, T5 and Flan-T5. The QAG model based on BART\textsubscript{LARGE} proves to be the best aligned with gold reference question and answers among most of the metrics. As with other QG experiments and NLP in general, the larger models prove more reliable.

\paragraph{Multilingual evaluation.} \autoref{tab:qaa-multilingual} shows the test results of three multilingual models (mBART, mT5\textsubscript{SMALL} and mT5\textsubscript{BASE}) in seven languages other than English, using their corresponding language-specific SQuAD-like datasets in QG-Bench for fine-tuning and evaluation.\footnote{The result of mBART in German is zero. Upon further inspection, we found that the fine-tuned answer extraction module did not learn properly, probably due to the limited size of the German dataset. T5 models, however, proved more reliable in this case.} In this evaluation, no single LM produces the best results across the board, yet QAG models based on mT5\textsubscript{SMALL} and mT5\textsubscript{BASE} are generally better than those based on mBART.

\begin{table}[t]
\centering
\scalebox{0.75}{
\begin{tabular}{@{}l@{\hspace{5pt}}c@{\hspace{8pt}}c@{}}
\toprule
Model & QAAligned BS & QAAligned MS \\ \midrule
BART\textsubscript{BASE}    & 92.8 / 93.0 / 92.8 & 64.2 / 64.1 / 64.5 \\
BART\textsubscript{LARGE}   & \textbf{93.2} / \textbf{93.4} / \textbf{93.1} & \textbf{64.8} / 64.6 / \textbf{65.0} \\
T5\textsubscript{SMALL}     & 92.3 / 92.5 / 92.1 & 63.8 / 63.8 / 63.9 \\
T5\textsubscript{BASE}      & 92.8 / 92.9 / 92.6 & 64.4 / 64.4 / 64.5 \\
T5\textsubscript{LARGE}     & 93.0 / 93.1 / 92.8 & 64.7 / \textbf{64.7} / 64.9 \\\midrule
Flan-T5\textsubscript{SMALL}& 92.3 / 92.1 / 92.5 & 63.8 / 63.8 / 63.8 \\
Flan-T5\textsubscript{BASE} & 92.6 / 92.5 / 92.8 & 64.3 / 64.4 / 64.3 \\
Flan-T5\textsubscript{LARGE}& 92.7 / 92.6 / 92.9 & 64.6 / \textbf{64.7} / 64.5 \\
\bottomrule
\end{tabular}
}
\caption{QAAligned scores ($F_1$/$P$/$R$) on the test set of SQuAD dataset by different QAG models, where the best score in each metric is shown in boldface.}
\label{tab:qaa-en}
\end{table}

\begin{table}[t]
\centering
\scalebox{0.75}{
\begin{tabular}{@{}llll@{}}
\toprule
 & Language & QAAligned BS   & QAAligned MS   \\\midrule
\multirow{7}{*}{\rotatebox{90}{mT5\textsubscript{SMALL}}}  
& German       & \textbf{81.2} / \textbf{80.0} / \textbf{82.5} & \textbf{54.3} / \textbf{54.0} / \textbf{54.6} \\
& Spanish      & 79.9 / 77.5 / 82.6 & 54.8 / 53.3 / 56.5 \\
& French       & \textbf{79.7} / \textbf{77.6} / \textbf{82.1} & \textbf{53.9} / \textbf{52.7} / \textbf{55.3} \\
& Italian      & 81.6 / 81.0 / \textbf{82.3} & \textbf{55.9} / 55.6 / \textbf{56.1} \\
& Japanese     & 79.8 / 76.8 / 83.1 & 55.9 / 53.8 / 58.2 \\
& Korean       & 80.5 / 77.6 / 83.8 & \textbf{83.0} / \textbf{79.4} / \textbf{87.0} \\
& Russian      & 77.0 / 73.4 / 81.1 & 55.5 / 53.2 / 58.3 \\\midrule
\multirow{7}{*}{\rotatebox{90}{mT5\textsubscript{BASE}}}  
& German       & 76.9 / 76.3 / 77.6 & 53.0 / 52.9 / 53.1 \\
& Spanish      & \textbf{80.8} / \textbf{78.5} / \textbf{83.3} & \textbf{55.3} / \textbf{53.7} / \textbf{57.0} \\
& French       & 68.6 / 67.6 / 69.7 & 47.9 / 47.4 / 48.4 \\
& Italian      & \textbf{81.7} / \textbf{81.3} / 82.2 & 55.8 / \textbf{55.7} / 56.0 \\
& Japanese     & \textbf{80.3} / \textbf{77.1} / \textbf{83.9} & \textbf{56.4} / \textbf{54.0} / \textbf{59.1} \\
& Korean       & 77.3 / 76.4 / 78.3 & 77.5 / 76.3 / 79.0 \\
& Russian      & 77.0 / 73.4 / 81.2 & 55.6 / 53.3 / 58.4 \\\midrule
\multirow{7}{*}{\rotatebox{90}{mBART}}  
& German       & 0 / 0 / 0          & 0 / 0 / 0          \\
& Spanish      & 79.3 / 76.8 / 82.0  & 54.7 / 53.2 / 56.4\\
& French       & 75.6 / 74.0 / 77.2 & 51.8 / 51.0 / 52.5 \\
& Italian      & 40.1 / 40.4 / 39.9 & 27.8 / 28.1 / 27.5 \\
& Japanese     & 76.7 / 74.8 / 78.9 & 53.6 / 52.3 / 55.1 \\
& Korean       & \textbf{80.6} / \textbf{77.7} / \textbf{84.0} & 82.7 / 79.0 / \textbf{87.0} \\
& Russian      & \textbf{79.1} / \textbf{75.9} / \textbf{82.9} & \textbf{56.3} / \textbf{54.0} / \textbf{58.9} \\\bottomrule
\end{tabular}
}
\caption{QAAligned scores ($F_1$/$P$/$R$) on the test set of QG-Bench by different QAG models, where the best score in each language is shown in boldface.}
\label{tab:qaa-multilingual}
\end{table}

\begin{table}[t]
\centering
\scalebox{0.75}{
\begin{tabular}{@{}l@{\hspace{5pt}}c@{\hspace{5pt}}c@{\hspace{5pt}}c@{\hspace{5pt}}c@{\hspace{5pt}}c@{\hspace{5pt}}c@{\hspace{5pt}}c@{}}\toprule
Gold & BART\textsubscript{B} & BART\textsubscript{L} & T5\textsubscript{S} & T5\textsubscript{B} & T5\textsubscript{L} & Flan-T5\textsubscript{S} & Flan-T5\textsubscript{B} \\\midrule
4.9  & 4.1                      & 4.2                       & 4.2                     & 4.3                    & 4.3                     & 4.2                          & 4.3                        \\
\bottomrule
\end{tabular}
}
\caption{Average number of generated question and answer pairs per paragraph on the test set of SQuAD by different QAG models.}\label{tab:size-en}
\end{table}

\begin{table}[t]
\centering
\scalebox{0.75}{
\begin{tabular}{@{}llccc@{}}
\toprule
Language & Gold & mT5\textsubscript{SMALL} & mT5\textsubscript{BASE} & mBART \\ \midrule
German   & 4.6  & 10.1                     & 8.4                     & 0.0   \\
Spanish  & 1.3  & 4.6                      & 4.8                     & 4.7   \\
French   & 1.3  & 4.9                      & 3.6                     & 5.4   \\
Italian  & 3.8  & 4.7                      & 4.6                     & 2.5   \\
Japanese & 1.3  & 6.6                      & 6.8                     & 3.6   \\
Korean   & 1.3  & 6.7                      & 6.3                     & 6.7   \\
Russian  & 1.3  & 4.8                      & 4.9                     & 4.7   \\
\bottomrule
\end{tabular}
}
\caption{The averaged number of generated question and answer pairs per paragraph on the test set of QG-Bench for each language.}\label{tab:size-multi}
\end{table}

\subsection{Number of Generated Questions and Answers}

\autoref{tab:size-en} and \autoref{tab:size-multi} show the averaged number of generated question-answer pairs and compare it to the number in the gold dataset. 
For English, there is a small difference across all QAG models, with all generating fewer pairs than the gold dataset, but with a limited margin.
For other languages, however, there are clear differences across QAG models, with the numbers of question-answer pairs generated by the QAG models always being larger than those in the gold dataset.
When comparing the number of pairs generated by the QAG models with their QAAligned scores, in languages such as German, Spanish, and Korean, QAG models that generated a larger number question-answer pairs achieved higher scores, not only recall-wise but also generally for F1.

\section{AutoQG}\label{sec:autoqg}

Finally, we present AutoQG (\url{https://autoqg.net}), an online QAG demo where users can generate question-answer pairs for texts in eight languages (English, German, Spanish, French, Italian, Japanese, Korean, Russian) by simply providing a context document.
We deploy the QAG models described in \autoref{sec:released-models}. In addition to the features described above, the online demo shows perplexity computed via \texttt{lmppl},\footnote{\url{https://pypi.org/project/lmppl}} a Python library to compute perplexity given any LM architecture. This feature helps us provide a ranked list of generation to the user.
Although we can compute perplexity for non-English generations based on the QAG models in each language, it entails large memory requirements on the the hosting server. As such, we  compute a lexical overlap between the question and the document as a computationally-light alternative to the perplexity, which is defined as:
\begin{align}
    1 - \frac{|q\cap p|}{|q|}
\end{align}
where $|\cdot|$ is the number of characters in a string, and $q\cap p$ is the longest sub-string of the question $q$ matched to the paragraph $p$.

\begin{figure}[t!]
 \centering
 \includegraphics[width=\columnwidth]{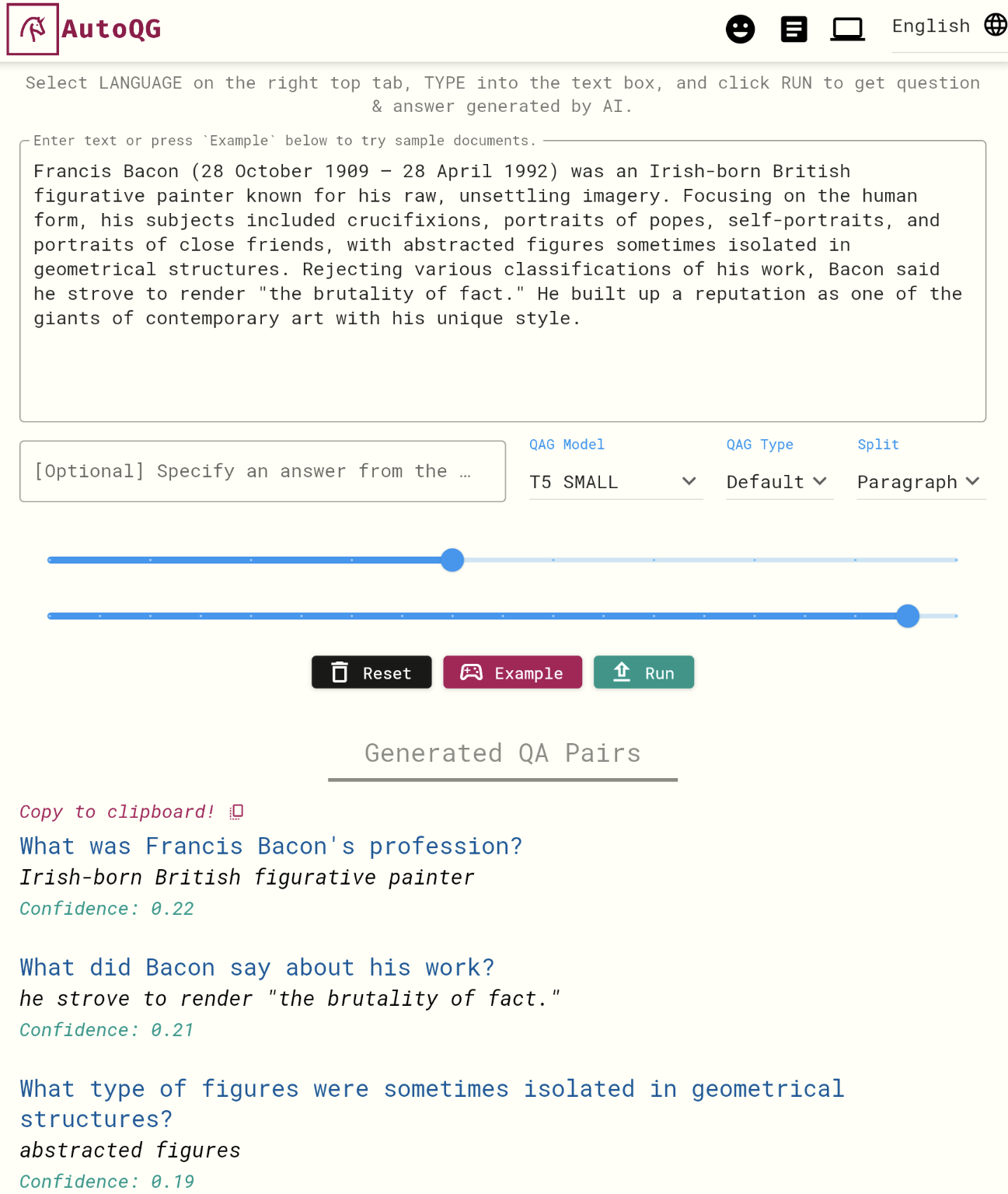}
\caption{A screenshot of AutoQG with an example of question and answer generation over a paragraph.}
 \label{fig:autoqg}
\end{figure}

\begin{figure}[t!]
 \centering
 \includegraphics[width=\columnwidth]{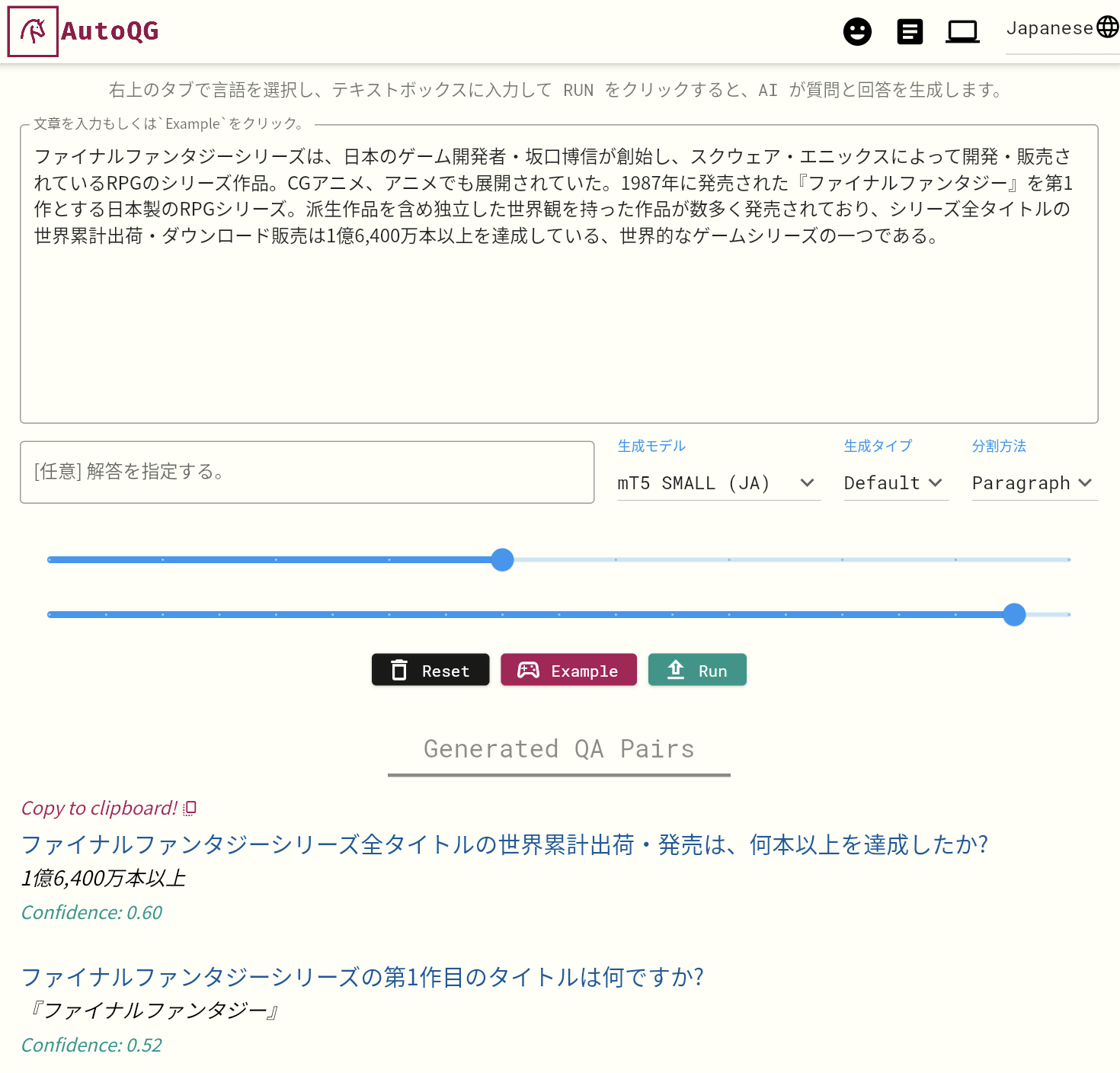}
\caption{A screenshot of AutoQG with an example of question and answer generation over a paragraph in Japanese.}
 \label{fig:autoqg-ja}
\end{figure}

\begin{figure}[t!]
 \centering
 \includegraphics[width=\columnwidth]{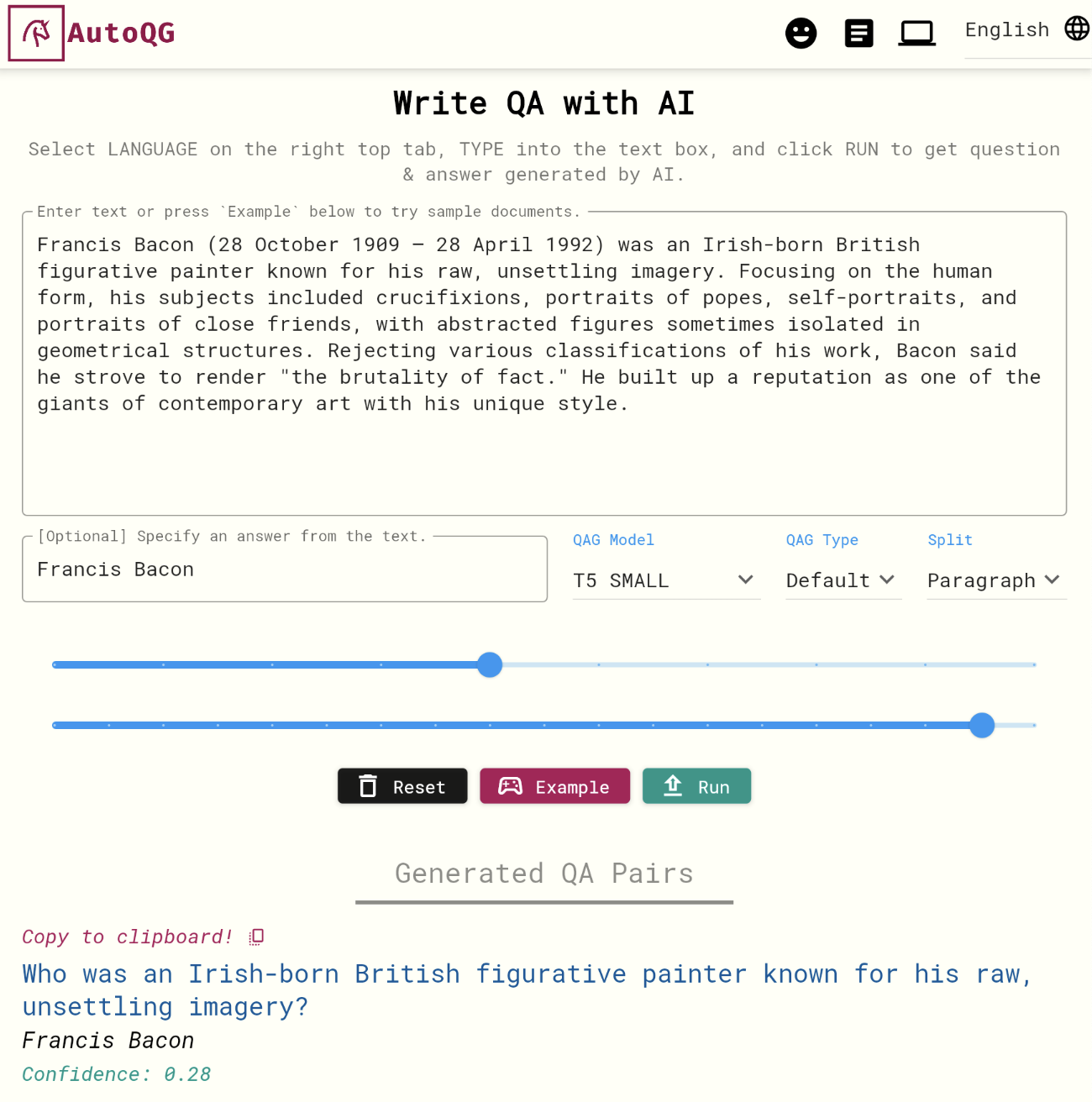}
\caption{A screenshot of AutoQG when an answer is specified by the user.}
 \label{fig:autoqg_answer}
\end{figure}

\autoref{fig:autoqg} and \autoref{fig:autoqg-ja} show examples of the interface with English and Japanese QAG, where there is a tab to select QAG models, language, and parameters at generation including the beam size and the value for nucleus sampling \cite{Holtzman2020The}.
Optionally, users can specify an answer and generate a single question on it with the QG model, as shown in \autoref{fig:autoqg_answer}.
A short introduction video to AutoQG is available at \url{https://youtu.be/T6G-D9JtYyc}.






\section{Conclusion}
In this paper, we introduced \texttt{lmqg}, a Python package to fine-tune, evaluate and deploy QAG models with a few lines of code. The library implements the QAG task as an efficient integration of answer extraction and question generation, and includes automatic reference-based metrics for model evaluation.
Finally, we showcase AutoQG, an online demo where end users can benefit from QAG models without any programming knowledge. AutoQG enables the selection of features going from different models and languages to controlling the diversity of the generation.

\section*{Limitations}
The focus on this paper was introducing software to make QAG models available to as many practitioners as possible, but there are a couple of limitations in the models and  evaluation metrics we proposed.

First, our released QAG models assume a paragraph up to around 500 tokens as an input, and longer documents can not be directly fed into the models. Additionally, the released QAG models were fine-tuned on questions that require one-hop reasoning only, so they are unable to generate multi-hop reasoning.

Second, the QAAligned score is a framework to extend any NLG metric to match the prediction to the reference when they are different in size, where we employed two well-established metrics (BERTScore and MoverScore) as underlying metrics. Since those underlying metrics are already proven to be effective \cite{zhang2019bertscore,zhao-etal-2019-moverscore,ushio-etal-2022-generative}, we have not conducted any human annotation for QAG specifically. Nonetheless, an extended human evaluation could help provide more insights on other limitations of the model not detected by the automatic evaluation.

\section*{Ethics Statement}
While the QAG models are fine-tuned on  pre-trained language models, which are known to contain some toxic contents \cite{schick-etal-2021-self}, an internal check does not reveal any toxic generation. However, there is a potential risk that the QAG model could generate toxic text due to the underlying LMs.

\bibliography{anthology,custom}

\begin{thebibliography}{34}
\expandafter\ifx\csname natexlab\endcsname\relax\def\natexlab#1{#1}\fi

\bibitem[{Bartolo et~al.(2021)Bartolo, Thrush, Jia, Riedel, Stenetorp, and
  Kiela}]{bartolo-etal-2021-improving}
Max Bartolo, Tristan Thrush, Robin Jia, Sebastian Riedel, Pontus Stenetorp, and
  Douwe Kiela. 2021.
\newblock \href {https://doi.org/10.18653/v1/2021.emnlp-main.696} {Improving
  question answering model robustness with synthetic adversarial data
  generation}.
\newblock In \emph{Proceedings of the 2021 Conference on Empirical Methods in
  Natural Language Processing}, pages 8830--8848, Online and Punta Cana,
  Dominican Republic. Association for Computational Linguistics.

\bibitem[{Bjerva et~al.(2020)Bjerva, Bhutani, Golshan, Tan, and
  Augenstein}]{bjerva-etal-2020-subjqa}
Johannes Bjerva, Nikita Bhutani, Behzad Golshan, Wang-Chiew Tan, and Isabelle
  Augenstein. 2020.
\newblock \href {https://doi.org/10.18653/v1/2020.emnlp-main.442} {{SubjQA}:
  {A} {D}ataset for {S}ubjectivity and {R}eview {C}omprehension}.
\newblock In \emph{Proceedings of the 2020 Conference on Empirical Methods in
  Natural Language Processing (EMNLP)}, pages 5480--5494, Online. Association
  for Computational Linguistics.

\bibitem[{Casimiro~Pio et~al.(2019)Casimiro~Pio, Marta~R., and Jose
  A.~R.}]{2016arXiv160605250R}
Carrino Casimiro~Pio, Costa-jussa Marta~R., and Fonollosa Jose A.~R. 2019.
\newblock \href {http://arxiv.org/abs/1912.05200v2} {{Automatic Spanish
  Translation of the SQuAD Dataset for Multilingual Question Answering}}.
\newblock \emph{arXiv e-prints}, page arXiv:1912.05200v1.

\bibitem[{Chung et~al.(2022)Chung, Hou, Longpre, Zoph, Tay, Fedus, Li, Wang,
  Dehghani, Brahma et~al.}]{chung2022scaling}
Hyung~Won Chung, Le~Hou, Shayne Longpre, Barret Zoph, Yi~Tay, William Fedus,
  Eric Li, Xuezhi Wang, Mostafa Dehghani, Siddhartha Brahma, et~al. 2022.
\newblock Scaling instruction-finetuned language models.
\newblock \emph{arXiv preprint arXiv:2210.11416}.

\bibitem[{Croce et~al.(2018)Croce, Zelenanska, and Basili}]{squad_it}
Danilo Croce, Alexandra Zelenanska, and Roberto Basili. 2018.
\newblock Neural learning for question answering in italian.
\newblock In \emph{AI*IA 2018 -- Advances in Artificial Intelligence}, pages
  389--402, Cham. Springer International Publishing.

\bibitem[{Denkowski and Lavie(2014)}]{denkowski-lavie-2014-meteor}
Michael Denkowski and Alon Lavie. 2014.
\newblock \href {https://doi.org/10.3115/v1/W14-3348} {Meteor universal:
  Language specific translation evaluation for any target language}.
\newblock In \emph{Proceedings of the Ninth Workshop on Statistical Machine
  Translation}, pages 376--380, Baltimore, Maryland, USA. Association for
  Computational Linguistics.

\bibitem[{d{'}Hoffschmidt et~al.(2020)d{'}Hoffschmidt, Belblidia, Heinrich,
  Brendl{\'e}, and Vidal}]{dhoffschmidt-etal-2020-fquad}
Martin d{'}Hoffschmidt, Wacim Belblidia, Quentin Heinrich, Tom Brendl{\'e}, and
  Maxime Vidal. 2020.
\newblock \href {https://doi.org/10.18653/v1/2020.findings-emnlp.107}
  {{FQ}u{AD}: {F}rench question answering dataset}.
\newblock In \emph{Findings of the Association for Computational Linguistics:
  EMNLP 2020}, pages 1193--1208, Online. Association for Computational
  Linguistics.

\bibitem[{Efimov et~al.(2020)Efimov, Chertok, Boytsov, and
  Braslavski}]{efimov2020sberquad}
Pavel Efimov, Andrey Chertok, Leonid Boytsov, and Pavel Braslavski. 2020.
\newblock Sberquad--russian reading comprehension dataset: Description and
  analysis.
\newblock In \emph{International Conference of the Cross-Language Evaluation
  Forum for European Languages}, pages 3--15. Springer.

\bibitem[{Heilman and Smith(2010)}]{heilman-smith-2010-good}
Michael Heilman and Noah~A. Smith. 2010.
\newblock \href {https://aclanthology.org/N10-1086} {Good question! statistical
  ranking for question generation}.
\newblock In \emph{Human Language Technologies: The 2010 Annual Conference of
  the North {A}merican Chapter of the Association for Computational
  Linguistics}, pages 609--617, Los Angeles, California. Association for
  Computational Linguistics.

\bibitem[{Holtzman et~al.(2020)Holtzman, Buys, Du, Forbes, and
  Choi}]{Holtzman2020The}
Ari Holtzman, Jan Buys, Li~Du, Maxwell Forbes, and Yejin Choi. 2020.
\newblock \href {https://openreview.net/forum?id=rygGQyrFvH} {The curious case
  of neural text degeneration}.
\newblock In \emph{International Conference on Learning Representations}.

\bibitem[{Lewis et~al.(2020)Lewis, Liu, Goyal, Ghazvininejad, Mohamed, Levy,
  Stoyanov, and Zettlemoyer}]{lewis-etal-2020-bart}
Mike Lewis, Yinhan Liu, Naman Goyal, Marjan Ghazvininejad, Abdelrahman Mohamed,
  Omer Levy, Veselin Stoyanov, and Luke Zettlemoyer. 2020.
\newblock \href {https://doi.org/10.18653/v1/2020.acl-main.703} {{BART}:
  Denoising sequence-to-sequence pre-training for natural language generation,
  translation, and comprehension}.
\newblock In \emph{Proceedings of the 58th Annual Meeting of the Association
  for Computational Linguistics}, pages 7871--7880, Online. Association for
  Computational Linguistics.

\bibitem[{Lewis et~al.(2019)Lewis, Denoyer, and
  Riedel}]{lewis-etal-2019-unsupervised}
Patrick Lewis, Ludovic Denoyer, and Sebastian Riedel. 2019.
\newblock \href {https://doi.org/10.18653/v1/P19-1484} {Unsupervised question
  answering by cloze translation}.
\newblock In \emph{Proceedings of the 57th Annual Meeting of the Association
  for Computational Linguistics}, pages 4896--4910, Florence, Italy.
  Association for Computational Linguistics.

\bibitem[{Lewis et~al.(2021)Lewis, Wu, Liu, Minervini, K{\"u}ttler, Piktus,
  Stenetorp, and Riedel}]{lewis-etal-2021-paq}
Patrick Lewis, Yuxiang Wu, Linqing Liu, Pasquale Minervini, Heinrich
  K{\"u}ttler, Aleksandra Piktus, Pontus Stenetorp, and Sebastian Riedel. 2021.
\newblock \href {https://doi.org/10.1162/tacl_a_00415} {{PAQ}: 65 million
  probably-asked questions and what you can do with them}.
\newblock \emph{Transactions of the Association for Computational Linguistics},
  9:1098--1115.

\bibitem[{Lim et~al.(2019)Lim, Kim, and Lee}]{lim2019korquad1}
Seungyoung Lim, Myungji Kim, and Jooyoul Lee. 2019.
\newblock Korquad1. 0: Korean qa dataset for machine reading comprehension.
\newblock \emph{arXiv preprint arXiv:1909.07005}.

\bibitem[{Lin(2004)}]{lin-2004-rouge}
Chin-Yew Lin. 2004.
\newblock \href {https://aclanthology.org/W04-1013} {{ROUGE}: A package for
  automatic evaluation of summaries}.
\newblock In \emph{Text Summarization Branches Out}, pages 74--81, Barcelona,
  Spain. Association for Computational Linguistics.

\bibitem[{Lindberg et~al.(2013)Lindberg, Popowich, Nesbit, and
  Winne}]{lindberg-etal-2013-generating}
David Lindberg, Fred Popowich, John Nesbit, and Phil Winne. 2013.
\newblock \href {https://aclanthology.org/W13-2114} {Generating natural
  language questions to support learning on-line}.
\newblock In \emph{Proceedings of the 14th {E}uropean Workshop on Natural
  Language Generation}, pages 105--114, Sofia, Bulgaria. Association for
  Computational Linguistics.

\bibitem[{Liu et~al.(2020)Liu, Gu, Goyal, Li, Edunov, Ghazvininejad, Lewis, and
  Zettlemoyer}]{liu-etal-2020-multilingual-denoising}
Yinhan Liu, Jiatao Gu, Naman Goyal, Xian Li, Sergey Edunov, Marjan
  Ghazvininejad, Mike Lewis, and Luke Zettlemoyer. 2020.
\newblock \href {https://doi.org/10.1162/tacl_a_00343} {Multilingual denoising
  pre-training for neural machine translation}.
\newblock \emph{Transactions of the Association for Computational Linguistics},
  8:726--742.

\bibitem[{Miller et~al.(2020)Miller, Krauth, Recht, and
  Schmidt}]{miller2020effect}
John Miller, Karl Krauth, Benjamin Recht, and Ludwig Schmidt. 2020.
\newblock The effect of natural distribution shift on question answering
  models.
\newblock In \emph{International Conference on Machine Learning}, pages
  6905--6916. PMLR.

\bibitem[{Möller et~al.(2021)Möller, Risch, and Pietsch}]{GermanQuAD}
Timo Möller, Julian Risch, and Malte Pietsch. 2021.
\newblock \href {http://arxiv.org/abs/2104.12741} {Germanquad and germandpr:
  Improving non-english question answering and passage retrieval}.

\bibitem[{Ousidhoum et~al.(2022)Ousidhoum, Yuan, and
  Vlachos}]{ousidhoum-etal-2022-varifocal}
Nedjma Ousidhoum, Zhangdie Yuan, and Andreas Vlachos. 2022.
\newblock \href {https://aclanthology.org/2022.emnlp-main.163} {Varifocal
  question generation for fact-checking}.
\newblock In \emph{Proceedings of the 2022 Conference on Empirical Methods in
  Natural Language Processing}, pages 2532--2544, Abu Dhabi, United Arab
  Emirates. Association for Computational Linguistics.

\bibitem[{Papineni et~al.(2002)Papineni, Roukos, Ward, and
  Zhu}]{papineni-etal-2002-bleu}
Kishore Papineni, Salim Roukos, Todd Ward, and Wei-Jing Zhu. 2002.
\newblock \href {https://doi.org/10.3115/1073083.1073135} {{B}leu: a method for
  automatic evaluation of machine translation}.
\newblock In \emph{Proceedings of the 40th Annual Meeting of the Association
  for Computational Linguistics}, pages 311--318, Philadelphia, Pennsylvania,
  USA. Association for Computational Linguistics.

\bibitem[{Puri et~al.(2020)Puri, Spring, Shoeybi, Patwary, and
  Catanzaro}]{puri-etal-2020-training}
Raul Puri, Ryan Spring, Mohammad Shoeybi, Mostofa Patwary, and Bryan Catanzaro.
  2020.
\newblock \href {https://doi.org/10.18653/v1/2020.emnlp-main.468} {Training
  question answering models from synthetic data}.
\newblock In \emph{Proceedings of the 2020 Conference on Empirical Methods in
  Natural Language Processing (EMNLP)}, pages 5811--5826, Online. Association
  for Computational Linguistics.

\bibitem[{Pyatkin et~al.(2021)Pyatkin, Roit, Michael, Goldberg, Tsarfaty, and
  Dagan}]{pyatkin-etal-2021-asking}
Valentina Pyatkin, Paul Roit, Julian Michael, Yoav Goldberg, Reut Tsarfaty, and
  Ido Dagan. 2021.
\newblock \href {https://doi.org/10.18653/v1/2021.emnlp-main.108} {Asking it
  all: Generating contextualized questions for any semantic role}.
\newblock In \emph{Proceedings of the 2021 Conference on Empirical Methods in
  Natural Language Processing}, pages 1429--1441, Online and Punta Cana,
  Dominican Republic. Association for Computational Linguistics.

\bibitem[{Raffel et~al.(2020)Raffel, Shazeer, Roberts, Lee, Narang, Matena,
  Zhou, Li, and Liu}]{T5}
Colin Raffel, Noam Shazeer, Adam Roberts, Katherine Lee, Sharan Narang, Michael
  Matena, Yanqi Zhou, Wei Li, and Peter~J Liu. 2020.
\newblock Exploring the limits of transfer learning with a unified text-to-text
  transformer.
\newblock \emph{Journal of Machine Learning Research}, 21:1--67.

\bibitem[{Rajpurkar et~al.(2016)Rajpurkar, Zhang, Lopyrev, and
  Liang}]{rajpurkar-etal-2016-squad}
Pranav Rajpurkar, Jian Zhang, Konstantin Lopyrev, and Percy Liang. 2016.
\newblock \href {https://doi.org/10.18653/v1/D16-1264} {{SQ}u{AD}: 100,000+
  questions for machine comprehension of text}.
\newblock In \emph{Proceedings of the 2016 Conference on Empirical Methods in
  Natural Language Processing}, pages 2383--2392, Austin, Texas. Association
  for Computational Linguistics.

\bibitem[{Schick et~al.(2021)Schick, Udupa, and
  Sch{\"u}tze}]{schick-etal-2021-self}
Timo Schick, Sahana Udupa, and Hinrich Sch{\"u}tze. 2021.
\newblock \href {https://doi.org/10.1162/tacl_a_00434} {Self-diagnosis and
  self-debiasing: A proposal for reducing corpus-based bias in {NLP}}.
\newblock \emph{Transactions of the Association for Computational Linguistics},
  9:1408--1424.

\bibitem[{So et~al.(2022)So, Byun, Kang, and Cho}]{so2022jaquad}
ByungHoon So, Kyuhong Byun, Kyungwon Kang, and Seongjin Cho. 2022.
\newblock Jaquad: Japanese question answering dataset for machine reading
  comprehension.
\newblock \emph{arXiv preprint arXiv:2202.01764}.

\bibitem[{Ushio et~al.(2022)Ushio, Alva-Manchego, and
  Camacho-Collados}]{ushio-etal-2022-generative}
Asahi Ushio, Fernando Alva-Manchego, and Jose Camacho-Collados. 2022.
\newblock {G}enerative language models for paragraph-level question generation.
\newblock In \emph{Proceedings of the 2022 Conference on Empirical Methods in
  Natural Language Processing}, Abu Dhabi, U.A.E. Association for Computational
  Linguistics.

\bibitem[{Ushio et~al.(2023)Ushio, Alva-Manchego, and
  Camacho-Collados}]{ushio-etal-2023-an-empirical}
Asahi Ushio, Fernando Alva-Manchego, and Jose Camacho-Collados. 2023.
\newblock An empirical comparison of lm-based question and answer generation
  methods.
\newblock In \emph{Proceedings of the 61th Annual Meeting of the Association
  for Computational Linguistics}, Toronto, Canada. Association for
  Computational Linguistics.

\bibitem[{Wolf et~al.(2020)Wolf, Debut, Sanh, Chaumond, Delangue, Moi, Cistac,
  Rault, Louf, Funtowicz, Davison, Shleifer, von Platen, Ma, Jernite, Plu, Xu,
  Le~Scao, Gugger, Drame, Lhoest, and Rush}]{wolf-etal-2020-transformers}
Thomas Wolf, Lysandre Debut, Victor Sanh, Julien Chaumond, Clement Delangue,
  Anthony Moi, Pierric Cistac, Tim Rault, Remi Louf, Morgan Funtowicz, Joe
  Davison, Sam Shleifer, Patrick von Platen, Clara Ma, Yacine Jernite, Julien
  Plu, Canwen Xu, Teven Le~Scao, Sylvain Gugger, Mariama Drame, Quentin Lhoest,
  and Alexander Rush. 2020.
\newblock \href {https://doi.org/10.18653/v1/2020.emnlp-demos.6} {Transformers:
  State-of-the-art natural language processing}.
\newblock In \emph{Proceedings of the 2020 Conference on Empirical Methods in
  Natural Language Processing: System Demonstrations}, pages 38--45, Online.
  Association for Computational Linguistics.

\bibitem[{Xue et~al.(2021)Xue, Constant, Roberts, Kale, Al-Rfou, Siddhant,
  Barua, and Raffel}]{xue-etal-2021-mt5}
Linting Xue, Noah Constant, Adam Roberts, Mihir Kale, Rami Al-Rfou, Aditya
  Siddhant, Aditya Barua, and Colin Raffel. 2021.
\newblock \href {https://doi.org/10.18653/v1/2021.naacl-main.41} {m{T}5: A
  massively multilingual pre-trained text-to-text transformer}.
\newblock In \emph{Proceedings of the 2021 Conference of the North American
  Chapter of the Association for Computational Linguistics: Human Language
  Technologies}, pages 483--498, Online. Association for Computational
  Linguistics.

\bibitem[{Zhang and Bansal(2019)}]{zhang-bansal-2019-addressing}
Shiyue Zhang and Mohit Bansal. 2019.
\newblock \href {https://doi.org/10.18653/v1/D19-1253} {Addressing semantic
  drift in question generation for semi-supervised question answering}.
\newblock In \emph{Proceedings of the 2019 Conference on Empirical Methods in
  Natural Language Processing and the 9th International Joint Conference on
  Natural Language Processing (EMNLP-IJCNLP)}, pages 2495--2509, Hong Kong,
  China. Association for Computational Linguistics.

\bibitem[{Zhang et~al.(2019)Zhang, Kishore, Wu, Weinberger, and
  Artzi}]{zhang2019bertscore}
Tianyi Zhang, Varsha Kishore, Felix Wu, Kilian~Q Weinberger, and Yoav Artzi.
  2019.
\newblock Bertscore: Evaluating text generation with bert.
\newblock In \emph{International Conference on Learning Representations}.

\bibitem[{Zhao et~al.(2019)Zhao, Peyrard, Liu, Gao, Meyer, and
  Eger}]{zhao-etal-2019-moverscore}
Wei Zhao, Maxime Peyrard, Fei Liu, Yang Gao, Christian~M. Meyer, and Steffen
  Eger. 2019.
\newblock \href {https://doi.org/10.18653/v1/D19-1053} {{M}over{S}core: Text
  generation evaluating with contextualized embeddings and earth mover
  distance}.
\newblock In \emph{Proceedings of the 2019 Conference on Empirical Methods in
  Natural Language Processing and the 9th International Joint Conference on
  Natural Language Processing (EMNLP-IJCNLP)}, pages 563--578, Hong Kong,
  China. Association for Computational Linguistics.

\end{thebibliography}
\bibliographystyle{acl_natbib}

\appendix

\section{Grid Search}
\label{sec:grid-search}

\begin{figure}[t!]
 \centering
 \includegraphics[width=0.65\columnwidth]{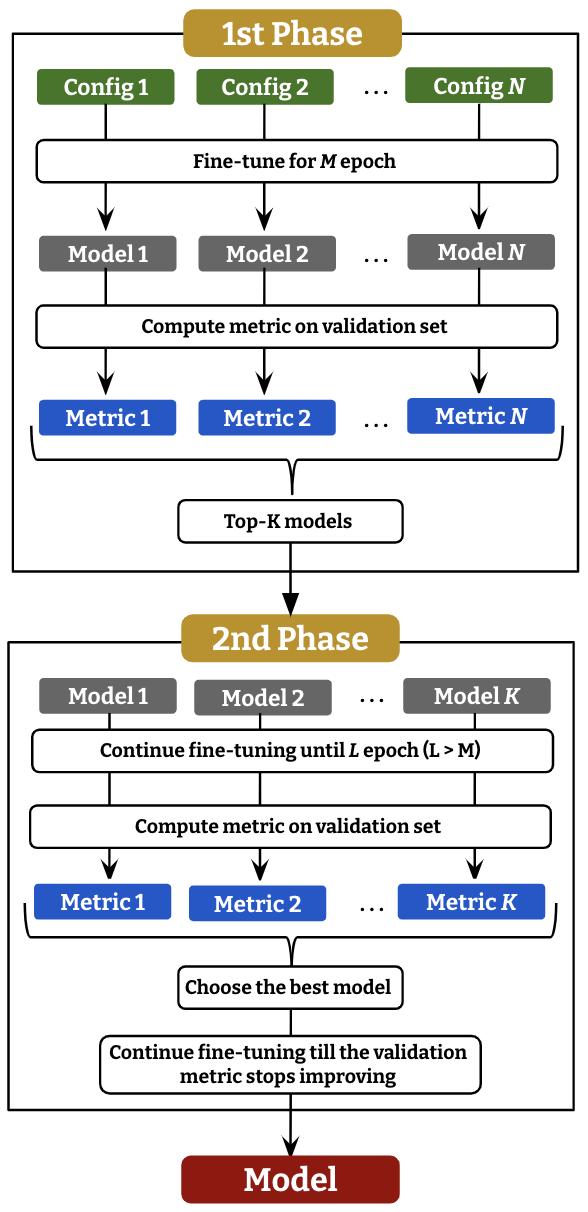}
\caption{An overview of the hyper-parameter search implemented as \texttt{GridSearcher}.}
 \label{fig:gridsearcher}
\end{figure}

To fine-tune LMs on QAG, one can use the \texttt{GridSearcher} class of \texttt{lmqg}, which performs LM fine-tuning with a two-stage optimization of hyper-parameter, a set of parameters to be used at fine-tuning such as learning rate or batch size, as described in \autoref{fig:gridsearcher}.
Let us assume that we want to find an optimal combination of the learning rate and random seed from a list of candidates [1e-4,1e-5] and [0,1] for learning rate and random seed respectively on QG as an example. 
We also assume a training and a validation dataset to train a model on the task and 
an evaluation score that reflects a performance of a model (eg. BLEU4\cite{papineni-etal-2002-bleu}),
and we define a search-space as a set including all the combinations of those candidates, i.e. \{(1e-4, 0), (1e-4, 1), (1e-5, 0), (1e-5, 1)\}.
The goal of the \texttt{GridSearcher} is to find the best combination to train a model on the training dataset for the target task over the search-space with respect to the evaluation score computed on the validation dataset.

Brute-force approach such as to train model over every combination in the search-space can be a highly-inefficient, so 
\texttt{GridSearcher} employs a two-stage search method to avoid training for all the combinations, while being able to reach to the optimal combination as possible.
To be precise, given an epoch size $L$ (\texttt{epoch}), the first stage fine-tunes all the combinations over the search-space, and pauses fine-tuning at epoch $M$ (\texttt{epoch\_partial}). The top-$K$ combinations (\texttt{n\_max\_config}) are then selected based on the evaluation score computed over the validation dataset, and they are resumed to be fine-tuned until the last epoch.
Once the $K$ chosen models are fine-tuned at second stage, the best model is selected based on the evaluation score, which is kept being fine-tuned until the evaluation score decreases.

The dataset for training and validation can be any datasets shared in the HuggingFace Hub, and one can specify the input and the output to the model from the column of the dataset by the arguments \texttt{input\_types} and \texttt{output\_types} at instantiating \texttt{GridSearcher}.
For example, the following code shows how we can fine-tune T5 \cite{T5} on question generation, a sub-task of QAG, with SQuAD \cite{rajpurkar-etal-2016-squad}, where the dataset \texttt{lmqg/qg\_squad} is shared at \url{https://huggingface.co/datasets/lmqg/qg_squad} on the HuggingFace Hub, which has columns of \texttt{paragraph\_answer}, that contains a answer-highlighted paragraph, and \texttt{question}, which is a question corresponding to the answer highlighted in the \texttt{paragraph\_answer}. We choose them as the input and the output to the model respectively by passing the name of each column to the arguments, \texttt{input\_types} and \texttt{output\_types}.
\begin{mypython}[label=SO-test]
  from lmqg import GridSearcher

  # instantiate the trainer
  trainer = GridSearcher(
    dataset_path="lmqg/qg_squad",
    input_types="paragraph_answer",
    output_types="question",
    model="t5-large",
    batch_size=128,
    epoch=10,
    epoch_partial=2,
    n_max_config=3,
    lr=[1e-4,1e-5],
    random_seed=[0,1])
  
  # train model
  trainer.train()
\end{mypython}

\end{document}